# Dimensionality Invariant Similarity Measure

Ahmad Basheer Hassanat

IT Department, Mutah University, Mutah – Karak, Jordan, 61710
ahmad.hassanat@gmail.com

**Abstract:** This paper presents a new similarity measure to be used for general tasks including supervised learning, which is represented by the K-nearest neighbor classifier (KNN). The proposed similarity measure is invariant to large differences in some dimensions in the feature space. The proposed metric is proved mathematically to be a metric. To test its viability for different applications, the KNN used the proposed metric for classifying test examples chosen from a number of real datasets. Compared to some other well known metrics, the experimental results show that the proposed metric is a promising distance measure for the KNN classifier with strong potential for a wide range of applications.

## 1. Introduction

A similarity measure is a function that gives a non-negative number to each pair of vectors to define a notion of likeness (Hagedoorn 2000). Such a measure is vital for a large number of applications and research areas including – but not limited to – pattern matching algorithms, artificial intelligence (Hagedoorn 2000), machine learning (He, Chen and Chen 2013), regression analysis (Wessela and Schork 2006) and data mining (Geng and Hamilton 2006), in addition to other research areas such as social science, economy, null theory testing etc. Similarity measures are needed in almost all knowledge disciplines.

A large number of similarity measures are proposed in the literature, perhaps the most famous and well known being the Euclidean distance stated by Euclid two thousand years ago. Over the last century great efforts have been made to find new metrics and similarity measures to satisfy the needs of different applications. New similarity measures are needed in particular for use in distance learning (Yang 2006), where classifiers such as the k-nearest neighbor (KNN) are heavily depended upon for choosing the best distance. Optimizing the distance metric is valuablein several computer vision tasks, such as object detection, content-based image retrieval, image segmentation and classification.

Cha has categorized similarity measures into eight families (Cha, 2007) and (Cha, 2008):
1- The Minkowski family, which includes Euclidean distance, City block or Manhattan distance and Chebyshev distance.
2- The absolute differencefamily, which includes Sørensen, Gower, Soergel, Kulczynski, Canberra and Lorentzian distances.
3- The intersection family, which includes Intersection, Czekanowski, Motyka, Kulczynski, Ruzicka and Tanimoto distances.
4- The inner product family, which includes Inner Product, Harmonic Mean, Cosine, Kumar-Hassebrook, Jaccard and Dice similarity measures.
5- The Fidelity family, which includes Fidelity similarity measure, Bhattacharyya, Hellinger, Matusita and Squared-chord distances.
6- Squared L2 family, which includes Squared Euclidean, Pearson, Neyman, Squared, Probabilistic Symmetric, Divergence, Clark and Additive Symmetric distances.
7- Shannon's entropy family, which includes Kullback–Leibler, Jeffreys, K divergence, Topsøe, Jensen-Shannon and Jensen difference distances.
8- Combinations (of the previous measure) family, which includes Taneja and Kumar-Johnson distances.

A much larger number of distances and similarity measures are illustrated in the work of (Deza and Deza 2009) showing the applications of each similarity measure. None of these functions measure similarity perfectly for all problems, as each deals with a specific data context and assumptions. According to the no free lunch theorem (Duda, Hart and Stork 2001), no distance function performs better than the other; the use of a particular distance is problem and data dependent.

The similarity measures that are used the most are Euclidean (ED) and Manhattan distances (MD)– both assume the same weight to all directions. In addition, the difference between vectors at each dimension might approach infinity to imply dissimilarity (Bharkad and Kokare 2011), therefore if there is abnormality in one value in any direction this will be reflected in the final result of the distance; for example, if we have two vectors of size 100,

V1=(1,2,3,...,99,100) and V2=(1,2,3,...,99,0), then ED(V1,V2) and MD(V1,V2) is 100. Perhaps both vectors are equal, but the last value is changed for some reason; noise, for instance, which shows that such distances are sensitive to large differences in any direction, and this allows some dimensions to dominate the distance even in the absence of noise.

To solve the previous problem, researchers usually opt for either normalization or standardization of the data. However, both have their own weaknesses. If there are outliers, most of the data will be forced to scale down using normalization; on the other hand, standardization degrades data and does not provide bounded data(Saitta 2007).

The so-called "*Wave-Hedges distance*" shown in Eq(1) solves part of the previous problem. This measure has been applied to *compressed image retrieval* (Hatzigiorgaki and Skodras 2003), *probability density function similarity* (Cha, 2007), *content based video retrieval* (Patel and Meshram 2012), *Image Retrieval* (Khapli and Bhalchandra 2011), (Braveen and Dhavachelvan 2009), *time series classification* (Giusti and Batista 2013), *landscape retrieval* (Jasiewicz, Netzel and Stepinski 2013), *image fidelity* (Macklem 2002), *Histogram Distance Measures* (Cha, 2008) and *finger print recognition* (Bharkad and Kokare 2011). Interestingly, the source of the "*Wave-Hedges*" metric has not been correctly cited, and some of the previously mentioned resources allude to it incorrectly as (Hedges 1976). The source of this metric eludes the author, despite best efforts otherwise.

Even the name of the distance "*Wave-Hedges*" is questioned, and therefore will not be used in the rest of this paper. Rather, we will refer to this distance as Eq(1) for the rest of this paper.

$$D_{WH}(A,B) = \sum_{i=1}^{m}\left(1 - \frac{min(A_i,B_i)}{max(A_i,B_i)}\right) \quad (1)$$

## 2. Material and Methods

There are 3 problems associated with Eq(1), those are:
1- It cannot deal correctly with points having 0 values. For example, if $A_i$ is equal to 0, then the distance between 0 and any other positive non-zero value is 1, no matter how large or small that value is.
2- The distance between 0 and 0 is undefined.
3- In addition, the distance is not well defined on points with negative values. For example, the distance between -1 and -2 is equal to -1, while the distance between 1 and 2 is 0.5.

This paper presents a new similarity measure based on Eq(1) to solve all the above-mentioned problems. The proposed similarity function between two points in two vectors is written as:

$$D(A_i, B_i) = \begin{cases} 1 - \frac{1+min(A_i,B_i)}{1+max(A_i,B_i)} & , min(A_i,B_i) \geq 0 \\ 1 - \frac{1+min(A_i,B_i)+|min(A_i,B_i)|}{1+max(A_i,B_i)+|min(A_i,B_i)|} & , min(A_i,B_i) < 0 \end{cases} \quad (2)$$

And for all the vectors dimensions we get:
$$D_{proposed}(A,B) = \sum_{i=1}^{m}\bigl(D(A_i,B_i)\bigr) \quad (3)$$

where A and B are both vectors with size m. $A_i$ and $B_i$ are real numbers.

As can be seen from Eq(2 and 3), the proposed measure is bounded by [0,1[. It reaches 1 when the maximum value approaches infinity assuming the minimum is finite, or when the minimum value approaches minus infinity assuming the maximum is finite. This is shown by Figure 1 and the following equations.

$$\lim_{max(A_i,B_i)\to\infty}\bigl(D(A_i,B_i)\bigr) = \lim_{min(A_i,B_i)\to-\infty}\bigl(D(A_i,B_i)\bigr) = 1 \quad (4)$$

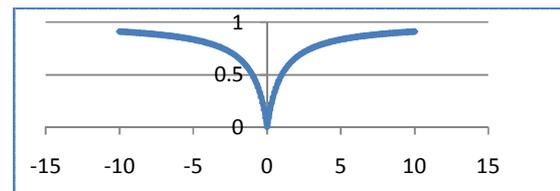

Figure 1. Representation of the proposed similarity measure between the point 0 and n, where n belongs to [-10, 10]

This means that the more a pair of values is similar, the nearest to zero the measure will be, and the more a pair of values is dissimilar, the nearest to one the measure will be. In other words, no matter what the difference between two values is, the distance will be in the range of 0 to 1.

By studying some properties of the proposed measure, such as non-negativity, equivalently, symmetry and Triangle inequality we may state the following Lemmas.

**Lemma 1**: the proposed similarity measure is non-negative function.

**Proof**: for any two positive numbers, the proposed distance uses the first part of Eq(1) so we need to prove that

$1 - \frac{1+min(A_i,B_i)}{1+max(A_i,B_i)} \geq 0$ This is equivalent to $\frac{max(A_i,B_i)-min(A_i,B_i)}{1+max(A_i,B_i)} \geq 0$ we get $\frac{|A_i-B_i|}{1+max(A_i,B_i)} \geq 0$

Since we assume two positive numbers then $max(A_i,B_i) \geq 0$, by dividing positive value $|A_i - B_i|$ by another positive value we get positive value, which is greater than 0.

If the minimum number is negative, the piecewise function uses the other formula which deals with negative numbers by adding the absolute value of the minimum number to both numerator and denominator. So we need to prove that

$$1 - \frac{1+min\,(A_i,B_i)+|mi\,n(A_i,B_i)|}{1+max\,(A_i,B_i)+|mi\,n(A_i,B_i)|} \geq 0$$

We know that $mi\,n(A_i, B_i) + |mi\,n(A_i, B_i)| = 0$ so we get

$$\frac{max\,(A_i,B_i)+|mi\,n(A_i,B_i)|}{1+max\,(A_i,B_i)+|mi\,n(A_i,B_i)|} \geq 0$$

We get $max\,(A_i, B_i) + |mi\,n(A_i, B_i)| \geq 0$
Now if $ma\,x(A_i, B_i) \geq 0$ the previous inequality is true. Else, we know that

$$|mi\,n(A_i, B_i)| \geq |max(A_i, B_i)|$$

By subtracting a smaller value ($-max\,(A_i, B_i)$) from a larger value $|mi\,n(A_i, B_i)|$ we get positive value, and this prove the previous inequality.

**Lemma 2**: the proposed similarity measure is an equivalence function.

**Proof**: a function D(A,B)=0 if and only if A coincides with B. If A coincides with B then the min(A,B)=max(A,B).
By dividing a value by itself we get 1. By subtracting this 1 from 1 as in our formula we get 0. We need to prove that:

$$1 - \frac{1+mi\,n(A_i,B_i)}{1+max(A_i,B_i)} = 0 \qquad (5)$$

By moving the negative part to the right we get:

$$1 = \frac{1+mi\,n(A_i,B_i)}{1+max(A_i,B_i)} \qquad (6)$$

If $A_i$ coincides with $B_i$ then:

$$mi\,n(A_i, B_i) = max(A_i, B_i) \qquad (7)$$

By replacing the *min* function with the *max* function we get:

$$1 = \frac{1+max(A_i,B_i)}{1+max(A_i,B_i)} \qquad (8)$$

And this gives:

$$1 = 1 \qquad \square \qquad (9)$$

Another approach for proving equivalently is to prove that D(A,A)=0, since we are considering the distance of a vector from itself. The min($A_i,A_i$)=max($A_i,A_i$)), by replacing the min with the max in Eq(5) we end up with 1-1=0.

**Lemma 3**: the proposed similarity function is symmetric.

**Proof**: a function D is symmetric if and only if D(A,B)= D(B,A) for all points A , B. so the proposed similarity measure should satisfy:

$$1 - \frac{1+mi\,n(A_i,B_i)}{1+max(A_i,B_i)} = 1 - \frac{1+mi\,n(B_i,A_i)}{1+max(B_i,A_i)} \qquad (10)$$

Here, we rely on the fact that the minimum and maximum functions are symmetric: min(A,B)=min(B,A) and max(A,B)=max(B,A). By substitution in the previous equation we get:

$$1 - \frac{1+mi\,n(A_i,B_i)}{1+max(A_i,B_i)} = 1 - \frac{1+mi\,n(A_i,B_i)}{1+max(A_i,B_i)} \qquad (11)$$

Moving the negative part to the right side of Eq(11) we get:

$$1 = 1 - \frac{1+mi\,n(A_i,B_i)}{1+max(A_i,B_i)} + \frac{1+mi\,n(A_i,B_i)}{1+max(A_i,B_i)} = 1 \qquad (12)$$

Both sides of the equation are equal, and then D is symmetric.

**Lemma 4**: the proposed similarity function satisfies the Triangle inequality.

**Proof**: a similarity function D satisfies the Triangle inequality if and only if D(A,C) ≤ D(A,B) + D(B,C) for all points A , B and C. so the proposed similarity measure should satisfy:

$$1 - \frac{1+mi\,n(A_i,C_i)}{1+max(A_i,C_i)} \leq 1 - \frac{1+mi\,n(A_i,B_i)}{1+max(A_i,B_i)} + 1 - \frac{1+mi\,n(B_i,C_i)}{1+max(B_i,C_i)} \qquad (13)$$

For simplicity, assume that ac=1+min(A,C), ab=1+min(A,B), bc=1+min(B,C), AC=1+max(A,C), AB=1+max(A,B) and BC=1+max(B,C) then Eq(13) becomes:

$$1 - \frac{ac}{AC} \leq 1 - \frac{ab}{AB} + 1 - \frac{bc}{BC} \qquad (14)$$

Moving the negative values to the opposite side of the equation we have:

$$1 + \frac{ab}{AB} + \frac{bc}{BC} \leq 2 + \frac{ac}{AC} \qquad (15)$$

Moving the 1 to the right we get:

$$\frac{ab}{AB} + \frac{bc}{BC} \leq 1 + \frac{ac}{AC} \qquad (16)$$

Notice that ac<= AC, because the minimum of two values is less than or equal to their maximum, therefore ac/AC<=1. Adding this inequality to the Eq(16) we get:

$$\frac{ab}{AB} + \frac{bc}{BC} + \frac{ac}{AC} \leq 2 + \frac{ac}{AC} \qquad (17)$$

By subtracting ac/AC from both sides we get:

$$\frac{ab}{AB} + \frac{bc}{BC} \leq 2 \qquad (18)$$

Notice that ab<=AB for the same previous reason, therefore:

$$\frac{ab}{AB} \leq 1 \qquad (19)$$

And bc<=BC, therefore:

$$\frac{bc}{BC} \leq 1 \qquad (20)$$

Combining the last two inequalities gives and proves Eq(18). $\square$

According to the previous discussion we can state the following theorem.

**Theorem 1**: the proposed similarity function is a metric.

**Proof**: a distance function should satisfy the following properties to be called a metric (Peeters, et al. 2008) and (Cha & Sriharib, 2002):
1- Non-negativity: D(A,B)>=0 for all points A , B.
2- Equivalently: D(A,B)=0 if and only if A coincides with B.
3- Symmetry: D(A,B)= D(B,A) for all points A , B.
4- Triangle inequality: D(A,C) <= D(A,B) + D(B,C) for all points A , B and C.

The previously proved Lemmas 1, 2, 3 and 4, show that the proposed distance function satisfies all the previous properties and therefore is a metric.

## 3. Results

For further study of the proposed metric, we opted for the KNN classifier, which naturally depends on similarity measures. We used only one neighbor 1-NN to ensure that any enhancement in the performance came from the similarity measure rather than the number of neighbors taken. For the experiments, we chose 19 different data sets from the UCI Machine Learning Repository (Bache and Lichman 2013). Table 1 depicts the data sets used.

Table 1. Description of the data sets used,

| Name | #Ex | #F | #C | data type | Min | Max |
|---|---|---|---|---|---|---|
| Heart | 270 | 25 | 2 | +Int | 0 | 564 |
| Balance | 625 | 4 | 3 | +Int | 1 | 5 |
| Cancer | 683 | 9 | 2 | +Int | 0 | 9 |
| German | 1000 | 24 | 2 | +Int | 0 | 184 |
| Liver | 345 | 6 | 2 | +Int | 0 | 297 |
| Vehicle | 846 | 18 | 4 | +Int | 0 | 1018 |
| Vote | 399 | 10 | 2 | +Int | 0 | 2 |
| Australian | 690 | 42 | 2 | + real | 0 | 100001 |
| Glass | 214 | 9 | 6 | + real | 0 | 75.41 |
| Sonar | 208 | 60 | 2 | + real | 0 | 1 |
| Wine | 178 | 13 | 3 | + real | 0.13 | 1680 |
| Diabetes | 768 | 8 | 2 | real & Int | 0 | 846 |
| Monkey1 | 556 | 17 | 2 | binary | 0 | 1 |
| Ionosphere | 351 | 34 | 2 | real | -1 | 1 |
| Phoneme | 5404 | 5 | 2 | real | -1.82 | 4.38 |
| Segmen | 2310 | 19 | 7 | real | -50 | 1386.33 |
| Vowel | 528 | 10 | 11 | real | -5.21 | 5.07 |
| Wave21 | 5000 | 21 | 3 | real | -4.2 | 9.06 |
| Wave40 | 5000 | 40 | 3 | real | -3.97 | 8.82 |

+Int: positive integer numbers
+real: positive real numbers
#Ex: number of examples
#F: number of features
#C: number of classes

Each data set is divided into two data sets– one for training and the other for testing. 30% of the data set is used for testing, and the rest of the data is for training. Each time the 1-NN is used to classify the test samples using ED, MD, Eq(1) and the proposed distance shown in Eq(3).

The 30% of data which is used as a test sample is chosen randomly, and each experiment on each dataset using a different distance is repeated 10 times to get random examples for testing and training. Table 2 shows the results of the experiments. The accuracy is averaged over the 10 runs.

Table 2. Comparison of the performance of the 1-NN using different distances

| Dataset | ED | MD | Eq1 | Proposed |
|---|---|---|---|---|
| Heart | 0.61 | 0.64 | 0.50 | **0.77** |
| Balance | 0.79 | 0.79 | **0.82** | **0.82** |
| Cancer | 0.95 | 0.96 | 0.77 | **0.96** |
| German | 0.66 | 0.68 | **0.71** | 0.69 |
| Liver | 0.59 | 0.60 | 0.61 | **0.62** |
| Vehicle | 0.63 | 0.67 | **0.67** | 0.66 |
| Vote | 0.92 | **0.93** | 0.49 | 0.92 |
| Australian | 0.65 | 0.69 | 0.55 | **0.82** |
| Glass | 0.70 | **0.71** | 0.36 | 0.67 |
| Sonar | 0.82 | 0.83 | 0.83 | **0.84** |
| Wine | 0.76 | 0.82 | 0.96 | **0.97** |
| Diabetes | 0.69 | **0.70** | 0.62 | 0.68 |
| Monkey1 | **0.79** | 0.79 | 0.48 | **0.79** |
| Ionosphere | 0.88 | 0.90 | 0.56 | **0.91** |
| Phoneme | 0.90 | 0.90 | 0.66 | **0.90** |
| Segmen | 0.96 | **0.97** | 0.14 | 0.96 |
| Vowel | 0.97 | **0.98** | 0.07 | 0.97 |
| Wave21 | **0.78** | 0.77 | 0.51 | 0.76 |
| Wave40 | **0.76** | 0.75 | 0.38 | 0.72 |
| Mean | 0.78 | 0.79 | 0.56 | **0.81** |

## 4. Discussions

As can be noticed from Table 2, the proposed distance achieved good results, outperforming the other distances in 10 datasets, and the overall average accuracy is the best among all the tested distances.

The lowest accuracies were recorded by Eq(1) as expected. This happens because of the problems mentioned earlier associated with Eq(1). For example, the accuracies of Eq(1) on *Vowel*, *Segmen* and *Waveform40* are 7.30%, 14.20% and 38.10% respectively. The reason for this low performance is because of the data type used on those datasets that contain real numbers (positive and negative values, see Table 1), while Eq(1) is not well defined in negative values.

Eq(1) also achieved low performance on other datasets, such as *Glass*, *Monkey1* and *Heart*, with accuracies of 35.80%, 47.50% and 49.80% respectively. This is justified by the number of zero values in those datasets. The data type of *Monkey1* is binary; this gives the zero values about a 50% chance of appearing in such a dataset. *Glass* and *Heart* also have a large number of zero values, This increases the probability of getting the distance from 0 to any number, which in this case will be 1 all the time, in addition to the "division by zero" problem which is also not defined in Eq(1).

On the contrary, we noted that Eq(1) performed very well on the *Wine* and *Balance* datasets. This is because all values there are non-negatives and non-zeros. It also performed well in *Sonar* and *German* datasets, where all values are positive with some zeros. These results justify the invention of the proposed metric Eq(3), which is inspired by Eq(1) and the modification made to fit all data types.

Interestingly, the Manhattan distance outperformed Euclidean distance in most datasets. This is because the difference between the paired values is squared in the ED, which emphasizes/reinforces the difference and allows one direction (feature) to dominate the result of the distance. This complies with some other researchers' results such as (Bonet, et al. 2008) and (Al Gindi, Attiatalla and Sami 2014), and contradicts others such as (Liu, et al. 2008). This reminds us again of the no-free-lunch theorem, i.e. there is no distance measure better than the other (including the proposed one) – it mainly depends on the problem and the data used.

This work proposes a new similarity measure, which we have proved mathematically as a metric function. This metric was compared to other well known metrics such as ED and MD in terms of accuracy.

Our results based on mathematical proofs and experiments on real data, show that the proposed metric is a promising distance measure, not only for the KNN classification but also for other problems and domains.

This complies with what other researchers have shown: that a well-defined distance metric may notably benefit KNN classifier performance compared to Euclidean distance (Yang 2006), (He, et al. 2004) and (Hastie and Tibshirani 1996).

Future work includes applying the proposed metric to other related problems such as k-mean clustering and hierarchal clustering to investigate its superiority in solving such problems.

**Acknowledgements:**
The author would like to acknowledge both Professor Christopher Farah of the University of Maine, and Fausto Galetto, ex-professor of Industrial Quality Management, Politecnico di Torino for their valuable discussions and comments that enriched this work.

**Corresponding Author:**
Dr. Ahmad Basheer Hassanat
IT Department
Mutah University,
Mutah – Karak, Jordan,
E-mail: ahmad.hassanat@gmail.com

8/5/2014